\documentclass[sigconf]{acmart}
\usepackage[section]{placeins}
\AtBeginDocument{%
  }

\setcopyright{acmlicensed}
\copyrightyear{2025}
\acmYear{2025}
\acmConference[KDD 2025 Workshop – Causal Inference and Machine Learning in Practice]
  {KDD 2025 Workshop – Causal Inference and Machine Learning in Practice}
  {August 4, 2025}
  {Toronto, ON, Canada}

\usepackage[utf8]{inputenc}
\DeclareUnicodeCharacter{202F}{\,}

\usepackage{graphicx}
\usepackage{booktabs}
\usepackage{amsmath,amsfonts}
\usepackage{multirow}
\usepackage{tabularx}
\usepackage{siunitx}
\usepackage{enumitem}
\usepackage[english]{babel}
\usepackage{microtype}

\usepackage{algorithm}
\usepackage{algpseudocode}

\floatname{algorithm}{\textbf{Algorithm}}

\algrenewcommand\algorithmicrequire{\textbf{Require:}}
\algrenewcommand\algorithmicensure{\textbf{Ensure:}}
\algrenewcommand\algorithmicprocedure{\textbf{Procedure:}}
\algrenewcommand\algorithmicfunction{\textbf{Function:}}
\algrenewcommand\algorithmicend{\textbf{end}}

\newcommand{\AlgShort}{\textbf{Alg.}\,}

\newcolumntype{Y}{>{\raggedright\arraybackslash}X}

\newcolumntype{Y}{>{\raggedright\arraybackslash}X}

\begin{document}

\title[SCM vs. Panel-Aware DML]{Dynamic Synthetic Controls vs. Panel-Aware Double Machine Learning for Geo-Level Marketing Impact Estimation}


\settopmatter{authorsperrow=3}  

\author[Lee et al.]{Sang Su Lee}
\affiliation{%
  \institution{Thumbtack, Inc.}
  \city{San Francisco}
  \state{CA}
  \country{USA}}
\email{psulee@thumbtack.com}

\author{Vineeth Loganathan}
\affiliation{%
  \institution{Thumbtack, Inc.}
  \city{San Francisco}
  \state{CA}
  \country{USA}}
\email{vloganathan@thumbtack.com}

\author{Vijay Raghavan}
\affiliation{%
  \institution{Thumbtack, Inc.}
  \city{San Francisco}
  \state{CA}
  \country{USA}}
\email{vraghavan@thumbtack.com}

\renewcommand{\shortauthors}{Lee et al.}

\begin{abstract}
Accurately quantifying geo-level marketing lift in two-sided marketplaces is challenging: the Synthetic Control Method (SCM) often exhibits high {\em power} yet systematically \emph{under-estimates} effect size, while panel-style Double Machine Learning (DML) is seldom benchmarked against SCM. We build an open, fully documented simulator that mimics a \emph{typical large-scale geo roll-out}: $N_{\text{unit}}$ regional markets are tracked for $T_{\text{pre}}$ weeks before launch and for a further $T_{\text{post}}$-week campaign window, allowing all key parameters to be varied by the user and probe both families under \textbf{five} stylised stress tests: (i) curved baseline trends, (ii) heterogeneous response lags, (iii) treated-biased shocks, (iv) a non-linear outcome link, and (v) a drifting control group trend.

Seven estimators are evaluated: three \textit{standard} Augmented SCM (ASC) variants and four panel-DML flavours (TWFE, CRE/Mundlak, first-difference, and within-group). Across 100 replications per scenario, \textbf{ASC models consistently demonstrate severe bias and near-zero coverage} in challenging scenarios involving nonlinearities or external shocks. By contrast, \textbf{panel-DML variants dramatically reduce this bias and restore nominal 95\%-CI coverage,} proving far more robust.

The results indicate that while ASC provides a simple baseline, it is \emph{unreliable} in common, complex situations. We therefore propose a \textbf{`diagnose-first' framework} where practitioners first identify the primary business challenge (e.g., nonlinear trends, response lags) and then select the \textbf{specific DML model best suited for that scenario}, providing a more robust and reliable blueprint for analysing geo-experiments.
\end{abstract}




\maketitle

\section{Introduction}
In two-sided marketplaces — such as ride-sharing platforms, home-sharing services, and e-commerce ecosystems — decision-makers are increasingly interested in understanding the causal impact of interventions on key business metrics. Whether it is a marketing campaign to boost user engagement or a subsidy to increase supply, the ability to reliably infer incremental gains ("lift") from such actions is crucial for optimal resource allocation. However, measuring incrementality in observational data is challenging due to confounding factors, temporal trends, and the complex dynamics inherent to marketplaces.

Causal machine learning has emerged as a promising toolset to tackle these challenges. Major platforms now integrate them into day-to-day decision making.  
Uber, for instance, built a spline-regularised learner that allocates marketing and incentive budgets across cities while explicitly accounting for causal lift~\cite{uber2024ciml}.  
Airbnb’s data-science group reports using causal inference to infer guest-demand elasticities and to optimise marketplace outcomes at scale~\cite{airbnb2024preferences}.  
Together, these cases illustrate a broader trend: firms are combining traditional econometric ideas with flexible ML models to answer causal questions at industrial scale.

Despite this progress, significant methodological questions remain. Traditional econometric approaches like diff-in-diff (DiD) and Synthetic Control Methods (SCM) have been go-to solutions for causal inference on aggregate, panel-structured data. SCM, in particular, has gained popularity for evaluating interventions in a single or small set of treated units by constructing a weighted synthetic comparator~\cite{abadie2010synthetic}. SCM accounts for time-varying confounders by matching pretreatment trends, which is a major advantage over DiD.

Double Machine Learning (DML), on the other hand, has emerged as a flexible ML-based framework to estimate treatment effects with complex confounders~\cite{chernozhukov2018double}. While DML has primarily been applied to cross-sectional data, recent advances propose adaptations for panel data. These adaptations allow DML to leverage time-invariant and time-varying covariates while addressing unobserved heterogeneity~\cite{clarke2024panel,fuhr2024rethinking}. Our paper explores how such panel-aware DML methods perform relative to SCM in a two-sided marketplace context.

We simulate weekly data for 200 geos over two years, designing realistic data-generating processes that incorporate nonlinear trends, heterogeneous treatment effects, biased external shocks, and nonlinear supply-demand matching. Across five scenarios, we evaluate how SCM and several panel-DML estimators compare in estimating average treatment effects. Our results provide practical guidance for analysts selecting among causal inference tools in complex real-world settings.

\section{Related Work}

Causal inference in panel settings has traditionally relied on diff-in-diff (DiD) and fixed effects models. Recent work has highlighted limitations of two-way fixed effects models under treatment heterogeneity or staggered adoption (e.g., Goodman-Bacon~\cite{goodman2021difference}; Sun and Abraham~\cite{sun2021estimating}). Alternative approaches such as doubly robust DiD~\cite{santanna2020doubly} and synthetic difference-in-differences~\cite{arkhangelsky2021synthetic} improve estimation by combining outcome modeling and propensity weighting.

Synthetic Control Methods (SCM), introduced by Abadie et al.~\cite{abadie2010synthetic}, construct weighted combinations of control units to approximate treated units’ counterfactual outcomes. SCM is widely used in policy evaluation and marketing analytics, particularly when a small number of treated units receive an intervention. Augmented SCM~\cite{ben2021augmented} further combines outcome modeling with SCM weighting for improved robustness.

Double Machine Learning (DML), proposed by Chernozhukov et al.~\cite{chernozhukov2018double}, estimates treatment effects using machine learning models for nuisance function estimation and orthogonalized causal regression. Recent extensions adapt DML to panel data using fixed effects, first-differencing, or correlated random effects transformations~\cite{clarke2024panel,fuhr2024rethinking}. These approaches enable flexible adjustment for observed and unobserved confounding.

In industry settings, causal–ML pipelines are now a standard part of
large-scale marketplace experimentation.
\citet{meta2022geolift} open-sourced \textsc{GeoLift}, a full
geo-experiment workflow that is widely used to quantify the offline
incrementality of online ad campaigns.
\citet{hermle2022valid} (LinkedIn Ads) propose an
‘‘asymmetric-budget-split’’ design and accompanying estimators that
deliver unobtrusive but statistically valid lift measurement at
nation-wide scale.
Our study complements these operational systems by benchmarking
Augmented SCM versus a family of panel-aware DML estimators under five
stress-test scenarios, thereby mapping recent methodological advances to
practical implementation choices that practitioners must make.

\section{Methodology}\label{sec:method}

This section formalises the causal estimand, summarises the two estimation paradigms under comparison, and specifies the evaluation metrics.

\subsection{Estimand}
Let $Y_{it}$ denote weekly gross revenue for geo $i$ in week $t$ ($i=1,\dots,N$, $t=0,\dots,T{-}1$).  $D_{it}\in\{0,1\}$ is an \textbf{active‑treatment indicator} that equals 1 only during the 12‑week exposure window of treated geos; $G_i=\max_t D_{it}$ is the \textbf{ever‑treated flag}.  Potential outcomes are $Y_{it}(1)$ and $Y_{it}(0)$.  We target the 
\emph{average treatment effect on the treated geos (ATT)} over the post‑period:


\begin{align}
\operatorname{ATT}
&= \frac{1}{N_T(T_{\mathrm{post}})}
     \sum_{i: G_i = 1}
     \sum_{t \in \mathcal{T}_{\mathrm{post}}}
     \bigl(Y_{it}(1) - Y_{it}(0)\bigr)
\label{eq:att}
\end{align}

\subsection{Synthetic Control Method}
\label{sec:method:scm}

\paragraph{Augmented Synthetic Control (ASC).}
Let $Y_{it}$ be the observed outcome for unit $i$ at time $t$,  
$\boldsymbol Y_{t}=[Y_{1t},\dots,Y_{Nt}]^{\top}$, and  
$D_{it}\in\{0,1\}$ the treatment indicator.  
Denote by $\mathcal T_0$ ($\mathcal T_1$) the pre- (post-) treatment periods and by  
$\mathcal I_\text{tr}$ ($\mathcal I_\text{don}$) the index sets of treated (donor) units.
ASC seeks weights $\boldsymbol w\in\mathbb R^{|\mathcal I_\text{don}|}$ that minimise


\begin{align}\label{eq:asc}
\min_{\boldsymbol w \ge 0,\;\mathbf{1}^\top \boldsymbol w = 1}
\;&
\sum_{t \in \mathcal{T}_0}
\bigl(Y_{it} - \sum_{j \in \mathcal{I}_{\mathrm{don}}} w_j\,Y_{jt}\bigr)^2
\;+\;
\lambda \lVert \boldsymbol w \rVert_2^2
\end{align}

where $\lambda$ is a ridge penalty controlling the weight dispersion.  
Given the fitted weights $\hat{\boldsymbol w}$, the counterfactual for each treated unit $i$ in
period $t>\min\mathcal T_1$ is


\begin{align}\label{eq:scm-att}
\widehat Y_{it}^{\mathrm{SCM}}
&= \sum_{j \in \mathcal{I}_{\mathrm{don}}} \hat w_j\,Y_{jt}, \\[6pt]
\widehat{\mathrm{ATT}}_t
&= \frac{1}{|\mathcal{I}_{\mathrm{tr}}|}
    \sum_{i \in \mathcal{I}_{\mathrm{tr}}}
      \bigl(Y_{it} - \widehat Y_{it}^{\mathrm{SCM}}\bigr)
\end{align}

We implement Equation~\eqref{eq:asc} with the \texttt{augsynth}  
package~\citep{benmichael2021augsynth}, using a ridge‐only prognostic function
and (unless otherwise noted) \textbf{no covariates} so that all methods are compared purely on their ability to exploit panel structure.

\vspace{0.5em}

\vspace{0.6em}
\noindent\textbf{Three ASC specifications.}
To diagnose the performance of synthetic controls, we evaluate three distinct specifications that incrementally enrich the feature space used to determine the donor weights:

\begin{enumerate}[label=(\alph*), leftmargin=*]
\item \textbf{ASC-Y} (``\emph{outcome-only}'' baseline) uses
      \emph{only} the past outcomes $Y_{it}$ when re-estimating the
      ridge–ASC weights.  
      It therefore assumes that pre-intervention trends are sufficient
      to screen off all latent differences between treated and donor
      markets.

\item \textbf{ASC-DEM} (``\emph{demographics}'') augments the
      outcome history with \emph{time-invariant covariates} so that the
      optimisation can explicitly balance structural demand
      fundamentals that are fixed within geo units but heterogeneous across
      space.  This mirrors the static-covariate term in our CRE-DML
      estimator.

\item \textbf{ASC-DEM-LAG} (``\emph{demographics + lagged demand}'')
      further adds \emph{lagged demand proxies}, most notably
      one- and two-week lagged search volume for the focal product
      category.  
      These pre-treatment lags capture high-frequency fluctuations in
      consumer interest that are \emph{predictive of near-future
      sales}, giving the synthetic control a chance to adjust for
      fast-moving demand shocks before the treatment starts.
\end{enumerate}

\smallskip
Putting the three specifications together lets us tease apart (i)~how much of the
\textbf{ASC's estimation error} stems from ignoring stable structural
differences (Y vs.\ DEM) and
(ii)~how much is due to not tracking short-run demand momentum
(DEM vs.\ DEM-LAG).
These variants appear in all subsequent tables under the abbreviations
given above.

\subsection{Panel‑Aware Double Machine Learning}\label{sec:method:pandml}
Our implementation follows the orthogonal‑residual recipe of Chernozhukov\,\textit{et~al.}\cite{chernozhukov2018double} but adapts each step to panel structure and high‑capacity learners written in \texttt{XGBoost}. \AlgShort\ref{alg:dml} summarizes the workflow.

\begin{algorithm}[h]
  \caption{\textbf{Cross‑fitted panel DML with IPTW and cluster SEs}}
  \label{alg:dml}
  \begin{algorithmic}[1]
    \State \textbf{Panel transformation:}
      Convert raw $(Y_{it},D_{it},X_{it})$ to
      $(Y^{\dagger},D^{\dagger},X^{\dagger})$ via one of:
      (i) TWFE dummies, (ii) geo‑demeaned (Within),
      (iii) first difference (FD), (iv) CRE/Mundlak.
    \State \textbf{Stratified geo folds:}
      Split rows by \emph{cluster‑balanced} cross‑fold so every fold
      contains treated and control geos.  For FD data ensure at least
      one $\Delta D\neq0$ per fold.
    \State \textbf{Nuisance learning:}
      \begin{itemize}[leftmargin=*,noitemsep]
        \item Outcome model: \texttt{XGBRegressor}
        \item Propensity model: \texttt{XGBClassifier}
      \end{itemize}
      Train on $\mathcal I_{\mathrm{train}}^{(k)}$ and predict on
      $\mathcal I_{\mathrm{test}}^{(k)}$ to get out‑of‑fold residuals
      $\hat\varepsilon_Y,\hat\varepsilon_D$.
    \State \textbf{IPTW stabilisation:}
      Compute
      \[
        w_i = \frac{D_i\,(1-\hat p_i)}{\hat p_i}
            + \frac{(1-D_i)\,\hat p_i}{1-\hat p_i}
      \]
      and trim the top 5 \% to avoid extreme weights.
    \State \textbf{Second‑stage WLS:}
      Regress $\hat\varepsilon_Y$ on $\hat\varepsilon_D$ with weights~$w$.
    \State \textbf{Uncertainty:}
      Report geo‑cluster robust SEs; optional unit‑bootstrap for coverage.
  \end{algorithmic}
\end{algorithm}

\paragraph{Why four variants?}\label{sec:method:why}
We include four panel transformations because each tackles a distinct threat to identification or efficiency.
\begin{enumerate}[label=(\alph*), leftmargin=*]
  \item \textbf{TWFE--DML (Two Way Fixed Effects; dummy absorption).} Adds $N{+}T{-}2$ dummies so the learner need not model unit or time intercepts. Best when $N$ and $T$ are modest and results must be comparable to a classical two–way fixed–effects regression.
  \item \textbf{WG--DML (Within-Group; geo demean).} Subtracts geo means before learning, $x_{it}^{w}=x_{it}-\bar x_i$. Algebraically identical to TWFE in linear settings yet far sparser, so boosting models avoid multicollinearity and memory blow-up when $N$ is large.
  \item \textbf{FD--DML (First Difference).} Uses $\Delta Y_{it}$ and $\Delta D_{it}$, wiping out every time-invariant component—including those correlated with $X_{it}$. The price is amplified measurement noise, but bias is minimal when unit-specific trends break strict exogeneity.
  \item \textbf{CRE--DML (Correlated Random Effect; Mundlak correction).} Augments $X_{it}$ with unit means $\bar X_i$ and treatment means $\bar D_i$, absorbing correlation between covariates and unobserved heterogeneity ($\mu_i$). Retains level information and often shows the best bias–variance trade-off when $T$ is moderate.
\end{enumerate}
These variants let us diagnose whether flexibility (CRE), noise attenuation (TWFE/WG), or non-stationarity robustness (FD) is most valuable under the scenarios in Section~\ref{sec:sim}.

\subsection{Implementation Details}\label{sec:method:impl}
All learners use identical hyper‑parameters across scenarios; no tuning leakage occurs because hyper‑parameters are fixed 
\emph{ex‑ante}.  Cross‑fitting splits by geo (not by time) to preserve within‑unit serial correlation.  SCM is implemented via the \texttt{augsynth} R package with ridge‑augmented option.


\section{Simulation Framework and Scenarios}\label{sec:sim}

\subsection{Full Data-Generating Process}

The simulation incorporates a rich set of features to model a realistic marketplace. Each geo is described by set of \textbf{time-invariant features}. The weekly dynamics are driven by \textbf{time-variant features} representing \textbf{demand features}, \textbf{supply features}, \textbf{competitor activity}, and \textbf{conversion} funnels that link demand to the final outcome.

\label{sec:dgp-full}

\begin{table}[h]
{\setlength\tabcolsep{3pt}%
\renewcommand{\arraystretch}{0.9}%
\footnotesize%
  \label{tab:key-const}
   \caption{Key generator parameters (defaults shown right).}
  \begin{tabular}{@{}llc@{}}
    \toprule
    Description & Symbol & Default \\
    \midrule
    \# geographies               & $N_{\text{unit}}$     & 200  \\
    \# treated geos              & $N_{\text{trt}}$      & 40   \\
    Pre‑period length (weeks)    & $T_{\text{pre}}$      & 52   \\
    Treatment window (weeks)     & $T_{\text{post}}$     & 12   \\
    Annual baseline growth       & $\mu_{\text{growth}}$ & 1.20 \\
    Peak proportional lift       & $\tau_{\max}$         & 0.23 \\
    Seasonality amplitude        & $A_{\text{season}}$   & 0.23 \\
    Noise s.d.                   & $\sigma_{\varepsilon}$& 0.10 \\
    Latent intercept s.d.        & $\sigma_{\eta}$       & 0.23 \\
    Weeks per season cycle       & $T_{\text{season}}$    & 52   \\
    \bottomrule
  \end{tabular}

}\end{table}


\paragraph{Step–by–step generation.}
For geo $i=1,\dots,N_{\text{unit}}$ and week  
$t = 1,\dots,T_{\text{pre}}+T_{\text{post}}$:

\begin{enumerate}[leftmargin=*]
\item \textbf{Baseline trend.}
\[
\log Y^{\mathrm{base}}_{it}
  = \alpha_i
    + \beta_i \frac{t}{T_{\mathrm{pre}}}
    + \gamma_i \sin\!\!\left(\frac{2\pi\,t}{T_{\mathrm{season}}}\right),
\]

with $\mathbb E[\beta_i]=\log(\mu_{\text{growth}})/52$ so the average
unit grows by $\mu_{\text{growth}}$ in one year.

\item \textbf{Stochastic noise.}\;
$\varepsilon_{it}\sim\mathcal N(0,\sigma_{\varepsilon}^2)$ and  
$Y^{0}_{it}= \exp\bigl(\log Y^{\text{base}}_{it}\!+\!\varepsilon_{it}\bigr)$.

\item \textbf{Treatment assignment.}\;
Randomly pick $N_{\text{trt}}$ units to form $\mathcal T$ and set  
$D_{it}=\mathbf 1\{i\in\mathcal T \land t>T_{\text{pre}}\}$.

\item \textbf{Lagged-impact curve.}
\[
\tau_{it}= \tau_{\max}\,\mathbf 1\{i\in\mathcal T\}
\bigl[\mathrm{sigmoid}\!\bigl(\tfrac{t-T_{\text{pre}}}{3}\bigr)
     -\mathrm{sigmoid}\!\bigl(\tfrac{t-T_{\text{pre}}-T_{\text{post}}}{3}\bigr)\bigr].
\]

\item \textbf{Observed outcome.}\;
$Y_{it}= Y^{0}_{it}\bigl(1+\tau_{it}\bigr)$, and we export the full
panel $\{(Y_{it},D_{it},X_{it})\}$.
\end{enumerate}

All constants in parentheses are easy to tweak for alternative stress
tests. Here $\eta_i$ introduces unobserved, unit-specific shifts in the
sigmoid-linked outcome model of Scenario S4, creating nonlinear
heterogeneity not captured by observed covariates.

\subsection{Stress-Test Scenarios}
\label{sec:sim:scenarios}

\begin{table}[h!]
 \caption{Stress–test scenarios. Each adds one failure mode to the base DGP.}
 \begin{tabularx}{\linewidth}{@{}lYl@{}}
    \toprule
    ID & Added complexity & Failure target \\
    \midrule
    S1 & Quadratic baseline trend: $\tau(t)=1+\alpha_1 t+\alpha_2 t^2$ & ASC extrapolation bias \\
    S2 & Geo-specific response lags/decays & Dynamic mis-specification bias \\
    S3 & Shock $+B_{\text{shock}}$ only in treated units & Hidden confounding \\
    S4 & Sigmoid outcome link with $\eta_i$ intercepts & Non-linear model mismatch \\
    S5 & Control group trend drift: $+\alpha_{drift}(t-T_{pre})$ & Parallel trends violation \\
    \bottomrule
 \end{tabularx}
 \label{tab:scenarios}
\end{table}

\vspace{0.5em}
\noindent\textbf{Stress-test scenarios.} 
Table\,\ref{tab:scenarios} shows the five perturbations we apply on top of the
baseline DGP (Section~\ref{sec:dgp-full}).  
Each is crafted to trigger a different, well-known failure mode of SCM- or
DML-style estimators.

\paragraph{S1: Non-linear baseline trend.}
We endow every unit with a small \emph{quadratic drift} =
$\log Y^{\text{base}}_{it}
  \;{+}\;\beta^{(2)}_i\!\bigl(t/T_{\text{pre}}\bigr)^{\!2}$,
where $\mathbb E[\beta^{(2)}_i]\!<\!0$.  
Because Ridge-regularised ASC relies on (approximately) linear
projection of donor trends,
curvature causes systematic \emph{under-extrapolation} and thus
downward-biased ATT.

\paragraph{S2: Heterogeneous response lags.}
Treatment effects now follow geo-specific impact curves  
$\tau_{it}= \tau_{\max}\ *f_i\!\bigl(t{-}T_{\text{pre}}\bigr)$
with randomly drawn onset, peak and decay parameters.
Static weights become mis-aligned with the moving effect window,
leading to \emph{dynamic-misspecification} bias.  
First-difference or de-meaned DML variants are expected to fare better.

\paragraph{S3: Shock larger in treated.}
We inject an exogenous post-period shock
$\delta_{\text{shock}}\!\sim\!\mathcal N(0,\sigma_{\text{shock}}^2)$
\emph{only} for units in $\mathcal T$.
ASC must separate the genuine treatment signal from this hidden
confounder; DML can mitigate bias if predictive covariates proxy
the shock mechanism.

\paragraph{S4: Non-linear outcome link.}
The linear growth term in the revenue equation is replaced by
$\eta_i\ *\mathrm{sigmoid}\!\bigl(\tfrac{t}{T_{\text{pre}}}-0.5\bigr)$,
inducing a strongly non-linear $X{\rightarrow}Y$ relationship.
This stresses outcome models that assume linearity and tests whether
flexible learners in panel-DML can adapt without mis-specification.

\paragraph{S5: Control group drift.} We add a scenario where the baseline trend for control units is modified during the post-treatment period. Specifically, the log-baseline outcome for control units is augmented by an additional term, $\mathbb{I}(i \notin \mathcal{I}_{tr} \land t \ge T_{\text{pre}}) \cdot \alpha_{\text{drift}} \cdot (t - T_{\text{pre}})$, where $\alpha_{\text{drift}} > 0$. This directly violates the parallel trends assumption, creating a situation where the control group is no longer a valid baseline and is likely to cause severe underestimation of the true treatment effect.

\vspace{0.5em}
Together these stress tests probe (i)~trend-extrapolation, (ii)~timing heterogeneity, (iii)~hidden confounding, (iv)~functional-form robustness, and (v)~robustness to baseline drift—dimensions where SCM and DML are known to exhibit complementary strengths and weaknesses.


\section{Experimental Results}  \label{sec:results}

\paragraph{Set-up.}
For each of the five stress scenarios, we generate $R=100$ independent panels, each containing $N_{\text{unit}}=200$ geo units of which $N_{\text{trt}}=40$ are randomly assigned to treatment after a $T_{\text{pre}}=52$-week pre-period, followed by a $T_{\text{post}}=12$-week intervention window. We then estimate the average treatment effect (ATT) on every replicate with the competing estimators. 

To evaluate performance, we focus on four key metrics: (i)~\textbf{Absolute Bias}, the average difference between the estimated and true effect (accuracy); (ii)~\textbf{Coverage}, the frequency with which the 95\% confidence interval contains the true effect (reliability); (iii)~\textbf{Power}, the frequency with which a true effect is correctly detected as statistically significant (sensitivity); and (iv)~\textbf{Avg. CI Width}, the average width of the confidence interval (precision).

\subsection{Performance in Scenario 1: Nonlinear Baseline Trend}
This scenario tests model performance against a quadratic baseline trend, a common feature of markets with accelerating growth. The results, shown in Table~\ref{tab:s1_results}, indicate that the linear assumptions of ASC models lead to severe under-estimation and high bias. In contrast, the flexible DML models adapt to the trend, with WG-DML providing the most accurate estimate (lowest bias) while maintaining high power.

\begin{table}[h!]
\centering
\small
\caption{Performance in Scenario 1: Nonlinear Baseline Trend}
\label{tab:s1_results}
\begin{tabular}{lrrrr}
\toprule
\textbf{Model} & \textbf{Abs. Bias} & \textbf{Coverage} & \textbf{Power} & \textbf{Avg. CI Width} \\
\midrule
ASC-Y & 5020.24 & 0.01 & 0.19 & 4415.34 \\
ASC-DEM & 5037.64 & 0.01 & 0.15 & 4547.66 \\
ASC-DEM-LAG & 5046.62 & 0.01 & 0.15 & 4631.11 \\
\midrule
CRE-DML & 4166.24 & 0.99 & 0.46 & 21385.11 \\
TWFE-DML & 2870.10 & 0.94 & 0.41 & 16366.87 \\
FD-DML & 2950.08 & 0.45 & 0.82 & 5527.35 \\
WG-DML & 1832.97 & 0.60 & 0.98 & 5785.75 \\
\bottomrule
\end{tabular}
\end{table}

\subsection{Performance in Scenario 2: Geo-Specific Response Lags}
This scenario introduces heterogeneous treatment lags, a challenge that resulted in critically low power for nearly all models, as shown in Table~\ref{tab:s2_results}. Although WG-DML offered the lowest bias and highest precision among DMLs, its 7\% power makes it unreliable for effect detection. In contrast, the structurally superior FD-DML model delivered the highest reliability (91\% coverage). Despite its own low power (2\%), its high reliability makes it the most trustworthy choice for this specific problem.

\begin{table}[h!]
\centering
\small
\caption{Performance in Scenario 2: Geo-Specific Response Lags}
\label{tab:s2_results}
\begin{tabular}{lrrrr}
\toprule
\textbf{Model} & \textbf{Abs. Bias} & \textbf{Coverage} & \textbf{Power} & \textbf{Avg. CI Width} \\
\midrule
ASC-Y & 269.80 & 1.00 & 0.01 & 4444.94 \\
ASC-DEM & 228.27 & 1.00 & 0.00 & 4423.62 \\
ASC-DEM-LAG & 224.97 & 1.00 & 0.00 & 4434.81 \\
\midrule
CRE-DML & 6372.29 & 0.94 & 0.06 & 426005.15 \\
TWFE-DML & 4584.69 & 0.90 & 0.10 & 416622.86 \\
FD-DML & 814.64 & 0.91 & 0.02 & 24187.34 \\
WG-DML & 744.03 & 0.67 & 0.07 & 12357.36 \\
\bottomrule
\end{tabular}
\end{table}

\subsection{Performance in Scenario 3: Treated-Only Shock}
Here, we test robustness to an external shock that positively impacts only the treated units. The results in Table~\ref{tab:s3_results} show that ASC models are unable to distinguish the shock from the treatment effect, leading to severe bias. DML models are more effective at isolating the true lift, with WG-DML providing the best balance of low bias, high power, and precision.

\begin{table}[h!]
\centering
\small
\caption{Performance in Scenario 3: Treated-Only Shock}
\label{tab:s3_results}
\begin{tabular}{lrrrr}
\toprule
\textbf{Model} & \textbf{Abs. Bias} & \textbf{Coverage} & \textbf{Power} & \textbf{Avg. CI Width} \\
\midrule
ASC-Y & 4973.75 & 0.01 & 0.19 & 4403.81 \\
ASC-DEM & 5000.56 & 0.01 & 0.15 & 4545.83 \\
ASC-DEM-LAG & 5010.39 & 0.01 & 0.14 & 4589.85 \\
\midrule
CRE-DML & 4130.50 & 0.99 & 0.44 & 621071.17 \\
TWFE-DML & 2811.69 & 0.94 & 0.47 & 716234.08 \\
FD-DML & 2928.82 & 0.41 & 0.83 & 5340.88 \\
WG-DML & 1822.77 & 0.63 & 0.97 & 55522.08 \\
\bottomrule
\end{tabular}
\end{table}

\subsection{Performance in Scenario 4: Nonlinear Outcome Link}
This scenario tests a complex, nonlinear relationship between inputs and outcomes, such as ad saturation. As seen in Table~\ref{tab:s4_results}, the DML models' flexibility is a clear advantage. WG-DML is again the top performer, delivering the lowest bias by a significant margin while maintaining high power.

\begin{table}[h!]
\centering
\small
\caption{Performance in Scenario 4: Nonlinear Outcome Link}
\label{tab:s4_results}
\begin{tabular}{lrrrr}
\toprule
\textbf{Model} & \textbf{Abs. Bias} & \textbf{Coverage} & \textbf{Power} & \textbf{Avg. CI Width} \\
\midrule
ASC-Y & 2722.90 & 0.03 & 0.17 & 2678.95 \\
ASC-DEM & 2713.44 & 0.02 & 0.11 & 2784.08 \\
ASC-DEM-LAG & 2719.71 & 0.03 & 0.11 & 2816.92 \\
\midrule
CRE-DML & 2749.44 & 1.00 & 0.33 & 514798.05 \\
TWFE-DML & 1894.94 & 0.95 & 0.30 & 211080.88 \\
FD-DML & 1593.13 & 0.54 & 0.76 & 93441.59 \\
WG-DML & 1046.38 & 0.69 & 0.95 & 83559.34 \\
\bottomrule
\end{tabular}
\end{table}

\subsection{Performance in Scenario 5: Control Group Anomaly}
This final scenario tests a critical failure of the parallel trends assumption, where the control group trend drifts. As shown in Table~\ref{tab:s5_results}, nearly all models failed, producing highly biased and unreliable estimates. The CRE-DML model was the sole exception. Despite having the lowest precision (the widest confidence interval), it was the only model to deliver both high reliability (98\% coverage) and the highest accuracy (lowest bias), proving it to be the most trustworthy choice for this severe challenge.

\begin{table}[h!]
\centering
\small
\caption{Performance in Scenario 5: Control Group Anomaly}
\label{tab:s5_results}
\begin{tabular}{lrrrr}
\toprule
\textbf{Model} & \textbf{Abs. Bias} & \textbf{Coverage} & \textbf{Power} & \textbf{Avg. CI Width} \\
\midrule
ASC-Y & 12936.77 & 0.00 & 0.00 & 4562.70 \\
ASC-DEM & 13307.00 & 0.00 & 0.00 & 4519.29 \\
ASC-DEM-LAG & 13442.66 & 0.00 & 0.00 & 4603.92 \\
\midrule
CRE-DML & 978.85 & 0.98 & 0.40 & 221408.58 \\
TWFE-DML & 2760.93 & 0.90 & 0.42 & 114984.97 \\
FD-DML & 2952.90 & 0.42 & 0.78 & 85453.30 \\
WG-DML & 3043.73 & 0.34 & 0.76 & 5931.74 \\
\bottomrule
\end{tabular}
\end{table}

\subsection{Summary of Key Findings}

The experimental results highlight the comparative strengths of DML and ASC models when faced with common, complex marketing dynamics. In our simulated environment, DML-based estimators consistently proved more robust than the standard ASC models, which showed significant bias and low reliability in scenarios with nonlinear trends, external shocks, or baseline drift.

However, these findings should be interpreted in the context of two important limitations of the simulation. First, our simulation does not include an expert-driven ``geo pre-selection'' step  (the process of manually selecting a more comparable set of control group markets before analysis), which in practice can significantly improve the performance of ASC by creating a more comparable control group. Second, the simulation does not explicitly model strong unobserved confounders, though scenarios like heterogeneous lags (S2) and external shocks (S3, S5) implicitly test the models' robustness to such challenges.

Considering these points, our findings do not suggest abandoning SCM, but rather adopting a more sophisticated, context-aware approach. The clear outperformance of specific DML models in each stress test points to a powerful practical takeaway: a \textbf{``diagnose-first'' strategy}. Analysts should first identify the most likely real-world challenge (e.g., nonlinear growth, potential for shocks) and then select the specific DML model best equipped to handle it. This approach, potentially used in parallel with a carefully curated SCM analysis, provides a robust and trustworthy framework for modern marketing analytics.

\subsection{Practical takeaway} \label{sec:practical-takeaways}

Our findings, which balance the simulation's results with its real-world limitations, lead to a set of actionable recommendations for practitioners. Instead of advocating for a wholesale replacement of one method with another, we recommend a more strategic, context-aware approach to geo-level impact analysis.

\begin{itemize}
    \item \textbf{Acknowledge Simulation Limitations in Practice:} Recognize that the raw performance of SCM in a simulation may not reflect its true performance in a business setting where expert ``geo pre-selection'' is applied. A carefully curated SCM can serve as a powerful and intuitive baseline.

    \item \textbf{Adopt a ``Diagnose-First'' DML Strategy:} The primary takeaway is to move from a ``one-size-fits-all'' model to a diagnostic approach. Before analysis, identify the most likely business challenge your campaign faces by referencing the five scenarios:
    \begin{itemize}
        \item For \textit{nonlinear trends or external shocks}, start with \textbf{WG-DML}.
        \item For \textit{suspected response lags}, use \textbf{FD-DML}.
        \item If you have concerns about \textit{unreliable control group trends}, \textbf{CRE-DML} is the most robust choice.
    \end{itemize}

    \item \textbf{Use Both Models for Robust Validation:} For high-stakes analyses, the most robust workflow is to run both a carefully curated SCM and the appropriate DML model in parallel. If their results align, confidence in the measured impact is high. If they diverge, it serves as a critical signal for a deeper investigation into potential unobserved confounders or model misspecifications.
    
    \item \textbf{Invest in Feature Engineering:} The success of DML models is highly dependent on a rich feature set. Continued investment in identifying and collecting data on potential business drivers is crucial for improving the accuracy and reliability of these advanced methods.
\end{itemize}

\section{Conclusion}
\label{sec:conclusion}

Our simulation provides a clear verdict on the comparative performance of Augmented Synthetic Controls and a suite of panel-aware Double Machine Learning models under several challenging, real-world scenarios. The results consistently demonstrated that while traditional ASC models provide an intuitive baseline, they are fragile. In the face of common complexities such as nonlinear trends, external shocks, and unreliable control groups, ASC models produced results with severe bias and near-zero coverage.

In contrast, DML models proved far more robust, consistently delivering more accurate and reliable estimates. Our key finding, however, is that no single DML model is universally superior. The optimal model is context-dependent: \textbf{WG-DML} excelled at handling nonlinearities and shocks, \textbf{FD-DML} was uniquely suited for heterogeneous response lags, and \textbf{CRE-DML} was the only effective tool against a drifting control group.

These findings should be contextualized by the realities of applied work. In practice, the performance of SCM is often enhanced by expert-driven \textbf{geo pre-selection}, and the success of DML is contingent upon rich \textbf{feature engineering} to make potential confounders observable. A purely model-vs-model comparison does not capture the critical role of this human-in-the-loop expertise.

Therefore, we conclude not with a declaration of a single ``winner,'' but with a recommendation for a more strategic workflow. We advocate for a \textbf{``diagnose-first'' approach}, where practitioners first identify the primary challenge of their specific geo-experiment and then select the DML model best equipped to handle it. This framework, which combines the analytical power of the right DML model with the crucial context provided by expert-driven data preparation, offers a more robust, reliable, and practical blueprint for modern marketing analytics. Future work should focus on empirically validating this hybrid, diagnostic workflow in a live production environment.


\appendix

\begin{table}[h]
\scriptsize
\setlength{\tabcolsep}{3pt}
    \caption{Frequently used symbols}
\begin{tabularx}{\linewidth}{@{}lY@{}}
    \toprule
    Symbol & Meaning \\
    \midrule
    \multicolumn{2}{@{}l}{\textit{Core Variables \& Estimand}} \\
    $Y_{it}$ & Outcome variable (weekly gross revenue) for unit $i$ at time $t$ \\
    $D_{it}$ & Treatment indicator for unit $i$ at time $t$ \\
    $ATT$    & Average Treatment effect on the Treated, the target estimand \\
    \midrule
    \multicolumn{2}{@{}l}{\textit{Simulation Parameters}} \\
    $N_{\text{unit}}$ & Total number of geo units in the simulation \\
    $T_{\text{pre}}$   & Length of the pre-treatment period (in weeks) \\
    $T_{\text{post}}$  & Length of the post-treatment (campaign) window \\
    $A_{\text{season}}$ & Amplitude of seasonal swing in baseline sales \\
    \midrule
    \multicolumn{2}{@{}l}{\textit{Scenario-Specific Parameters}} \\
    $B_{\text{shock}}$ & Additive shock applied to treated geos (S3) \\
    $\eta_i$           & Latent geo intercept for the non-linear term in Scenario 4 (S4) \\
    $\alpha_{\text{drift}}$ & Coefficient for the time-trend drift in Scenario 5 (S5) \\
    \midrule
    \multicolumn{2}{@{}l}{\textit{Model-Specific Parameters}} \\
    $\mathcal{I}_{tr}$ & The index set of treated (treatment group) units \\
    $w_i$              & Stabilised IPTW weight (Alg.~\ref{alg:dml}) \\
    \bottomrule
\end{tabularx}
\label{tab:glossary}

\end{table}

\FloatBarrier
\bibliographystyle{ACM-Reference-Format}

\end{document}